\def\corrEmail{yozkanay@nd.edu}

\documentclass[12pt]{article}
\usepackage{graphicx} 
\usepackage[margin=0.85in]{geometry}
\usepackage{datetime}
\usepackage{graphicx}
\usepackage{wrapfig}
\usepackage{hyperref}
\usepackage{lipsum}
\usepackage{lmodern}
\usepackage{picinpar}
\usepackage[dvipsnames]{xcolor}
\usepackage{tcolorbox}
\usepackage[export]{adjustbox}
\usepackage{lipsum}
\usepackage[font={small}]{caption}
\usepackage{adjustbox}
\makeatletter
\renewcommand{\maketitle}{\bgroup\setlength{\parindent}{0pt}
\begin{flushleft}
  \textbf{\@title}

  \@author
\end{flushleft}\egroup
}
\makeatother

\begin{document}

\begin{center}
\textbf{\large{The Effect of Tail Stiffness on a Sprawling Quadruped Locomotion}} \\
\vspace{5mm}
Josh Buckley\,$^{1}$, and Yasemin Ozkan-Aydin\,$^{2,*}$\\
{$^{1}$Department of Biomedical Engineering, University of Galway, Ireland \\
$^{2}$Department of Electrical Engineering, University of Notre Dame, IN, USA  }
\end{center} 

\begin{abstract}
\section*{}
A distinctive feature of quadrupeds that is integral to their locomotion is the tail. Tails serve many purposes in biological systems including propulsion, counterbalance, and stabilization while walking, running, climbing, or jumping. Similarly, tails in legged robots may augment the stability and maneuverability of legged robots by providing an additional point of contact with the ground. However, in the field of terrestrial bio-inspired legged robotics, the tail is often ignored because of the difficulties in design and control. This study will test the hypothesis that a variable stiffness robotic tail can improve the performance of a sprawling quadruped robot by enhancing its stability and maneuverability in various environments.  To test our hypothesis, we add a multi-segment, cable-driven, flexible tail, whose stiffness is controlled by a single servo motor in conjunction with a reel and cable system, to the underactuated sprawling quadruped robot. By controlling the stiffness of the tail, we have shown that the stability of locomotion on rough terrain and the climbing ability of the robot are improved compared to the movement with a rigid tail and no tail. The flexible tail design also provides passively controlled tail undulation capabilities through the robot's lateral movement, which contributes to stability.

\tiny
 \small{ \section*{Keywords:} quadruped locomotion, bio-inspired robotics, flexible tail, stability} 
\end{abstract}

\section{Introduction}
\begin{figure}[t]
    \centering
    \includegraphics[width=0.8\textwidth]{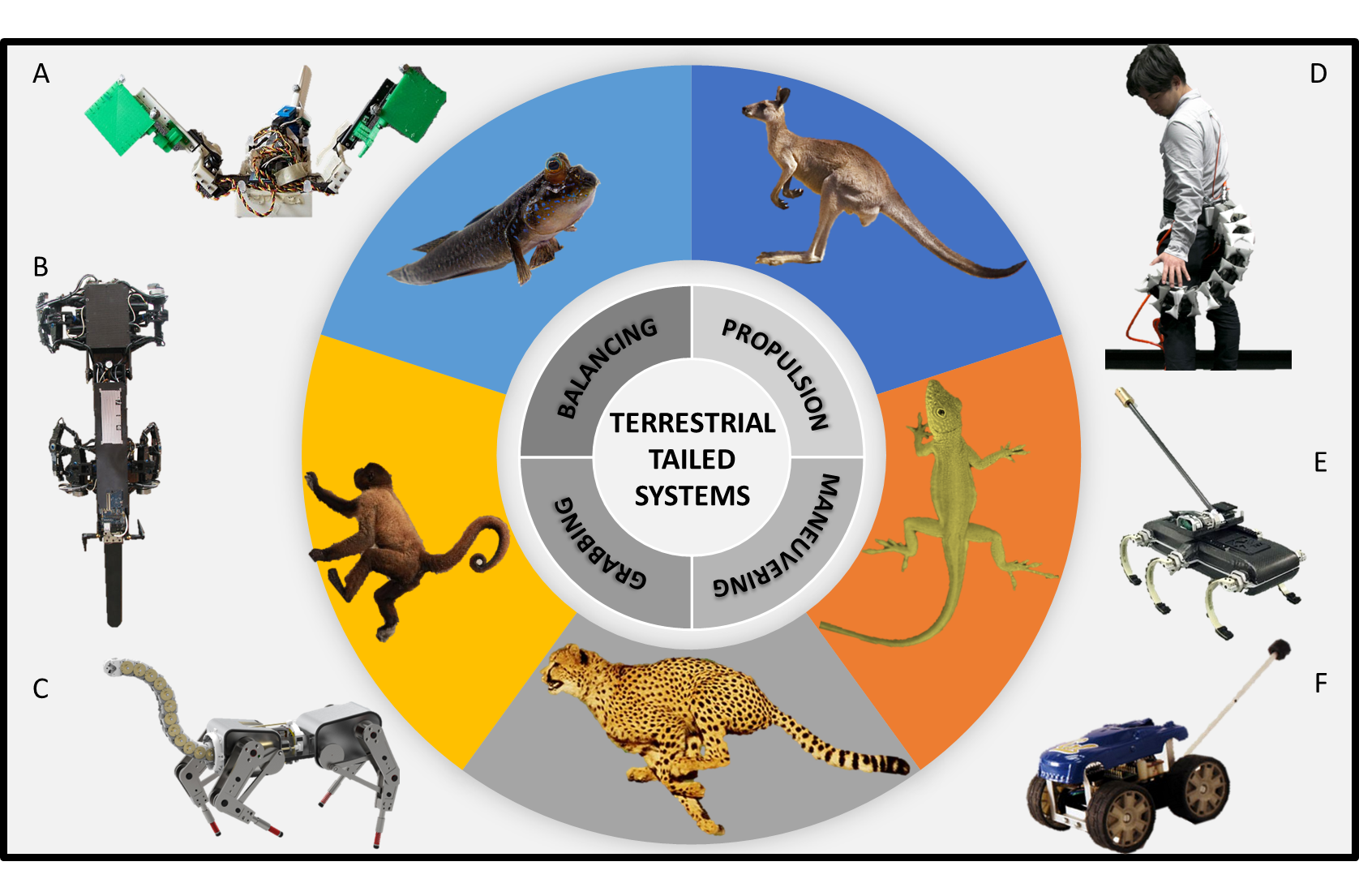}
    \caption{\textbf{Examples of terrestrial-tailed animals and robots.} The robotic systems depicted in the figure were designed with inspiration drawn from the anatomical and behavioral characteristics of the animals featured in the central image. \textbf{(A)} Muddybot inspired by mudskipper fish to demonstrate the function of a tail on the locomotion of granular surfaces \cite{muddybot2016}, \textbf{(B)} RISE Ver.3 climbing robot \cite{RISEV3}, \textbf{(C)} Articulated-tailed quadruped robot \cite{TailQuadruped2021}, \textbf{(D)} Prosthetic tail to extend human body functions \cite{nabeshima2019prosthetic}, \textbf{(E)} Tailed hexapod robot \cite{johnson2012tail},  \textbf{(F)} Tailbot, a lizard-inspired tailed quadruped robot \cite{chang2011lizard}.}
    \label{fig:TerrestrialTailedSystems}
\end{figure}

Tails have been observed to serve a multitude of functions in various extinct and extant animal species, ranging from mating and defense to aiding in locomotion \cite{MammalTailReview,ibrahim2020tail,santori2005locomotion,libby2012tail,muddybot2016}. Some animals, such as lizards, are capable of detaching their tails as a defensive mechanism to distract predators \cite{arnold1984evolutionary,higham2013integrative,naya2007some}. Tails also aid in maintaining balance \cite{MammalTailReview,walker1998balance} and stability \cite{TailStability}, especially during climbing or navigating through dense vegetation, and provide maneuvering \cite{gillis2016consequences} and grasping \cite{garber1999ecological} abilities. For example, many species of monkeys use their prehensile tail to grasp branches to support themselves while feeding \cite{PrehensileTail, CapachinTail}. 

In many cases, tails serve as a crucial propulsion system in both aquatic \cite{lauder2005hydrodynamics,lighthill1969hydromechanics,feilich2015passive,flammang2011volumetric} and terrestrial environments\cite{MammalTailReview,ibrahim2020tail,santori2005locomotion,muddybot2016}. Fish use their tails to move through the water by undulating antagonistic muscles that control tail oscillation \cite{lauder2005hydrodynamics}. Additionally, fish use their tails for steering and maintaining balance, making quick movements  and turning to avoid obstacles or predators \cite{keenleyside2012diversity}. Similarly, the tail assists in locomotion activities of terrestrial animals such as quadrupedal walking \cite{karakasiliotis2009analysis,willey2004tail,higham2013integrative}, climbing \cite{chang2011lizard}, jumping \cite{b2013impact,howell1944speed}, aerial descent \cite{young2015tail}, and gliding\cite{siddall2021mechanisms}. Some legged animals, such as kangaroos use their tails as a counterbalance during bipedal hopping, but when moving slowly, during grazing and social interactions. Animals use the tail as a ``fifth leg’’, providing an extra point of contact with the ground and an additional propulsive force \cite{kangaroo2014,karakasiliotis2009analysis}. 

These natural tail functionalities have served as a source of insight and inspiration for the development of novel robotic systems that emulate and replicate the natural movements and functionalities of the biological organisms \cite{TailQuadruped2021,nabeshima2019prosthetic}. The primary focus of the research conducted to date has been with regard to tails that allow for inertial adjustments to provide balance and stability \cite{CheetahRobotStabilityTail,Norby2021}, or maneuverability when turning \cite{CheetahTailTurning, QuadrupedRobotTurning,johnson2012tail,Casarez2017,CheetahTailTurning,jusufi2010righting}, running \cite{chang2011lizard}, jumping \cite{Brill2015} and climbing \cite{RISEV3,Lynch2012}. There has been limited research with respect to tails providing stability via continuous ground support, and tails that directly aid in propulsion. In \cite{TailTapRobot}, a tail was designed that periodically ``taps'' off the ground, providing a momentary ground contact. This design provided some success in obstacle navigation and releasing the robot when trapped.

Although there are several potential advantages of using tail-like appendages on robots, there are challenges that need to be overcome to fully realize the potential of tailed systems \cite{liu2021dynamic,Schwaner2021}. For example, one of the challenges in developing bio-inspired robotic tails is designing them to be flexible and agile enough to provide the desired benefits without adding too much weight or complexity to the overall system. In addition, we need to consider how to integrate the tail into the overall control system of the robot and how to power it. There is also a need for more sophisticated control algorithms and sensors that can accurately sense and respond to the movements and forces acting on the tail during locomotion. 

\begin{figure}[t]
    \centering
    \includegraphics[width=0.8\textwidth]{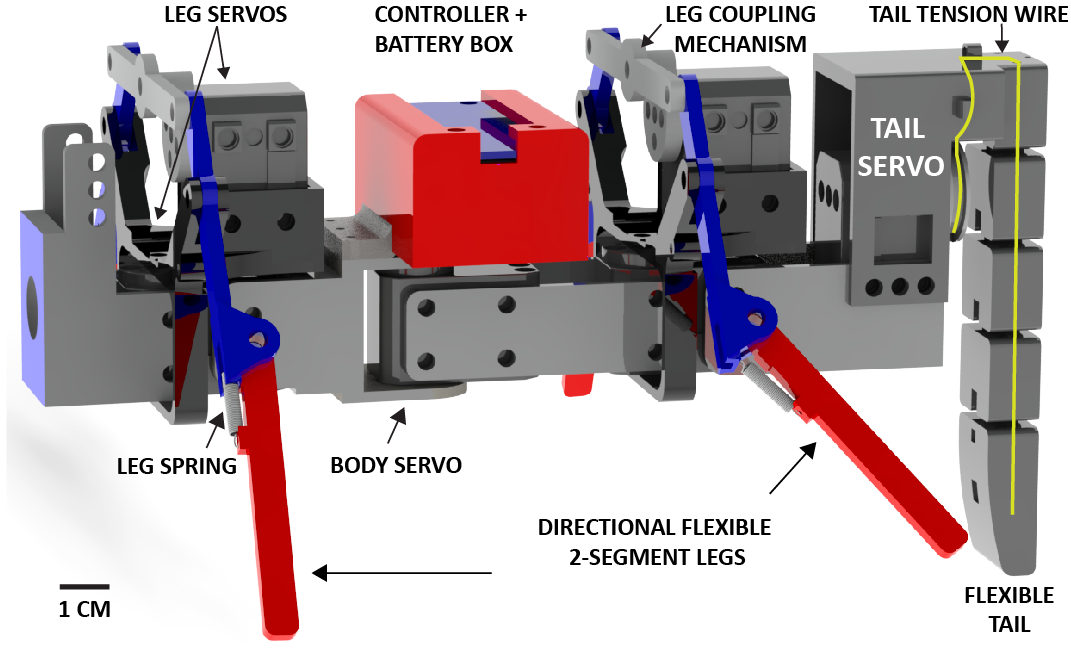}
    \caption{\textbf{A sprawling quadruped robot combined with a multi-segment flexible tail.} A computer-aided design of the fully autonomous quadruped robot  (about 20 cm long) that includes two body segments with a pair of directional flexible legs connected via an XL-320 servo motor. A 7.4 V LiPo battery and Robotis Open-CM 9.04 controller are placed into the box on the top of the body servo (red box). Two XL-320 servos control the horizontal and vertical motions of the coupled legs in a segment. A flexible, multi-segmented, cable-driven tail is attached to the back of the robot.    }
    \label{fig:RobotFigure_Labelled}
\end{figure}
In this paper, we study how the flexibility of the tail can enhance the locomotion performance of a sprawled quadruped robot used in \cite{OzkanAydin_Swarm2021} in various environments. We designed a 3D-printed, underactuated articulated tail whose stiffness could be controlled through a cable-driven servo-actuated system. Our tail design was inspired by the s-shaped tail of kangaroos, which is composed of several caudal vertebrae and provides great control when hopping and walking, as well as physical support when grazing \cite{kangaroo2014}. During the prototyping stages of the tail, it was found that an excess of vertebra-like segments led to an overabundance of flexibility. Subsequently, the reduced number of such segments is sufficient for emulating the characteristic behavior of a biological vertebrate tail. This approach allowed us to strike a balance between stability and flexibility, which was imperative for enhancing the robot's climbing capability. In evaluating the efficacy of our approach, we subjected our quadruped robot with a flexible tail to various environments.  Specifically, we tested the robot's performance on smooth surfaces, inclined and obstacle environments, and natural terrains. To provide a baseline, we compared the performance of the robot with that of the tailless and rigid-tailed robots in each of these environments.

\section{Materials and Methods}
\subsection{Quadruped Robot Design}
In this study, we added a multisegmented tail to the quadruped robot that was used in the previous studies \cite{OzkanAydin2020Centipede,OzkanAydin_Swarm2021}. The robot has two body segments each consisting of two directional flexible multi-jointed legs (total length = 12 cm) that are coupled via a rigid 1-DoF mechanism (Fig.\ref{fig:RobotFigure_Labelled}). Springs (spring constant = 0.2 kg/cm, McMaster; product number, 9654K949) that are connected between the upper and lower part of the leg joints allow the legs to bend when encountering obstacles and then return to a neutral position passively once the obstacle has been traversed \cite{OzkanAydin2020Centipede}.When the legs are at their neutral angle (i.e. the leg coupling mechanism is parallel to the ground), the height of the CoM of the robot is 5 cm. All the robot parts were 3D-printed with a Stratasys F170 printer using ABS material. 

The robot consists of a Robotis OpenCM 9.04 microcontroller, a lithium polymer battery (11.1 V, 1000 mAh), and six Robotis Dynamixel XL-320 motors (stall torque is 0.39 N.m): four for controlling the vertical and horizontal motion of the coupled legs, one for controlling the rotation of the body, and one for controlling tail stiffness. Because of the mechanical constraints, in the horizontal plane, the legs can rotate from $-25^o$ to $30^o$, and the body servo can rotate from $-30^o$ to $30^o$ from their neutral positions. The servos that control the vertical motion of the legs can rotate $30^o$, which raises the tip of the leg about 4 cm above the ground.

\subsection{Tail Design}
\subsubsection{Rigid Tail}
The rigid tail was designed to be of similar shape to the robot's legs (rectangular), but longer in size (90 mm in length, 3.5 mm thick, and 8 mm wide, Fig. \ref{fig:flexibleTailStates}A). The top of the tail was connected to the rear servo motor. The bottom of the tail was filleted, with the removal of sharp corners key to ensuring the tail does not get stuck on obstacles as it moves. The primary goal of this design was to provide an extra contact point at the back section of the robot. Since the legs are coupled, only one of the legs of each segment are on the ground at each time step \cite{OzkanAydin_Swarm2021}, and therefore cannot create a support triangle. Without a tail, the diagonal gait of the robot was interrupted and led to instability during locomotion \cite{OzkanAydin_Swarm2021}, with the body of the tail striking the ground twice per gait cycle. This also led to foot slippage and insufficient ground clearance, which leads to problems when the robot attempts to climb and navigate obstacles.

This tail acted passively, but its lateral motion could also be controlled by the rear servo motor which allows the tail to lift off the ground. This mechanism could be useful in applications such as climbing or walking on rough terrain. Were the robot to get stuck on an obstacle, repeated rotation of the rigid tail from side to side could push the robot out of this position.

\subsubsection{Flexible Tail - Mechanism and Actuation Design}

The tail was comprised of five articulating vertebra-like segments of cuboidal shape (Fig. \ref{fig:RobotFigure_Labelled}). The fifth segment of the tail has a narrow, curved end, which was in contact with the ground. This segment included a fillet on the interior face, allowing for the tail to pass over obstacles and sharp ledges, without causing the robot to get stuck. The tail was 3D printed using a Stratasys F170 printer with all parts intact. The links between each segment were designed in order to eliminate the need for assembly. Dimensions of the linkage component can be modified in order to alter the rigidity of the tail along with the range of motion.

Included in the design was a connection component, a box attached to the back of the robot. This component allowed the tail to attach and detach from the robot with ease. It also raised the attachment point of the tail to a point much higher than the body. This was important as it allowed a longer tail to be developed, leading to an increased range of motion, and thus more effective robot locomotion. The connection component also included small loops at the top and at the side, which helped guide the cabling when wrapping around the reel.

\begin{figure}
    \centering
    \includegraphics[width=1\textwidth]{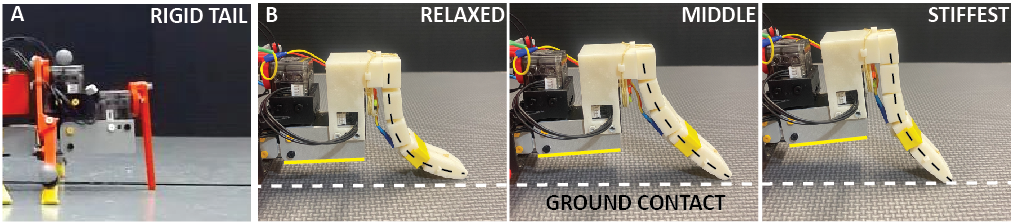}
    \caption{\textbf{Rigid and Flexible Tail. A.} Rigid tail \textbf{B.} States of the flexible tail. Tail stiffness is increasing from left to right, causing the height of the rear of the body to increase (SI Movie). }
    \label{fig:flexibleTailStates}
\end{figure}

To control the flexibility of the tail we designed a mechanism that consists of a custom-made reel that is attached to the rear servo motor (Fig. \ref{fig:RobotFigure_Labelled} and  Fig. \ref{fig:flexibleTailStates}). Kevlar wire was then threaded through the front and rear channels of the tail and wrapped around the reel. which was attached to the servo motor. The function of the rear wire was to flex the tail inwards (towards the body), while the front wire had the opposite function; it extends/raises the tail upwards (away from the body). These two cables can be configured to act as an antagonistic pair as follows. The two cables are wrapped around the reel in opposite directions (rear cable in an anticlockwise direction, front cable in a clockwise direction). When the servo motor is in its initial condition, the tail is relaxed and very flexible. By rotating the servo motor anticlockwise, the tension in the rear cable was increased and the front cable was relaxed. This tension resulted in a force that drove the tail toward the body. As the ground was obstructing the tail's motion, the tail applied a force on the ground which resulted in pushing the robot higher. It was noted that this could be beneficial in climbing applications.

A different approach to this antagonistic pair configuration is to increase the rotation range of the tail, allowing the tail to completely flex towards the body. The front surface of the tail (surface furthest from the body) is then in contact with the ground. The servo motor would then rotate in a clockwise direction, essentially 'flicking' the tail in the opposite direction and extending the tail away from the body. This approach resulted in a greater generated force and drove the body of the robot higher. This flicking mechanism could be implemented periodically to help stop the robot from getting stuck on ledges or objects.

Both of these approaches resulted in driving the robot upwards, increasing the clearance height of the legs and improving climbing ability. However, these designs had an adverse effect on the robot's stability. To solve this, the cabling was reconfigured so that both cables were wrapped around the reel in the same direction (anticlockwise). The tension in the cables was carefully tuned so that when the reel rotated anticlockwise, the tail would be pulled into an approximate 's' shape (Fig.\ref{fig:flexibleTailStates}-B), similar to the shape of vertebrate tails observed in certain biological species such as the kangaroo \cite{kangaroo2014}. With an applied rotation of the tail servo motor of between 55 to 90 degrees (anticlockwise), the distance between the segments was reduced, which increases the rigidity (or decreases the flexibility) of the tail.   To return the tail to its fully relaxed, flexible state, the servo motor was rotated in a clockwise direction. Tension in the cables would be released and would have no effect on the motion of the tail, allowing it to act passively. Two different implementations of this approach were tested, primarily the effect of periodically stiffening and relaxing the tail during key phases of the gait cycle, and the effect of using a touch sensor to sense when the tail should stiffen and relax. More detail on the stiffening/relaxation timing can be found in the Results Section 3.3.3.

\section{Experimental Results}
The effectiveness of the controlled flexible tail on the robot's locomotion was tested and compared to identical tests carried out on the robot with a rigid tail and the robot with no tail. The experiments aimed to simulate diverse and challenging terrains that the robot might encounter in real-world scenarios. Four scenarios were chosen to test the robot's locomotion capabilities; \textbf{(1)} walking on a flat surface (indoor), \textbf{(2)} climbing a series of steps (indoor), \textbf{(3)} walking on inclined surfaces (indoor), and \textbf{(4)} traversal of rough, natural terrain (outdoor). For each indoor experiment, there were five trials completed. The robot began at the same location for each trial, and the body position was kept constant by returning the servo motors to their starting position before each test was conducted. In all of the experiments, the robot walked with a diagonal gait in which the diagonal legs are on the ground at 50\% of the stance phase while undulating its body, \cite{OzkanAydin_Swarm2021,OzkanAydin2020Centipede}. The experiments were captured using a side view and a vertical aerial view using two Logitech C920 pro webcams.

In order to investigate the effectiveness of the flexible tail in facilitating stable and adaptive locomotion in natural environments, a series of outdoor experiments were conducted on diverse terrains including mulch, pebbles, and sand. The experiments were designed to demonstrate the challenging and unpredictable characteristics of natural environments and to evaluate the robot's ability to maintain balance and stability on such surfaces. 

\begin{figure}[t!]
    \centering
    \includegraphics[width=0.8\textwidth]{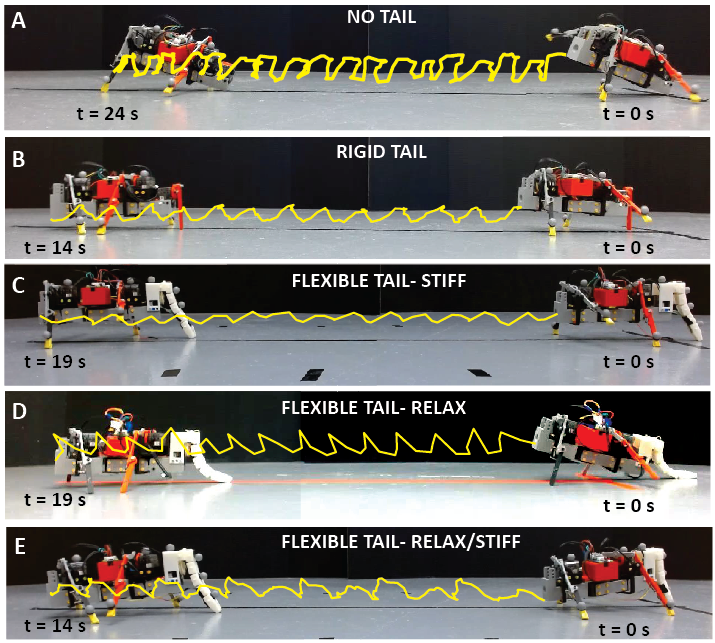}
    \caption{\textbf{Locomotion on flat terrain with different tail configurations.} Experimental snapshots of the robot from side view with \textbf{A} no tail \textbf{B.} a rigid tail \textbf{C.} a flexible tail with a stiff setting, \textbf{D.}  the flexible tail with a relaxed setting, E. the flexible tail with an alternating relax/stiff setting. The yellow trajectories show the tip trajectory of the robots during walking (SI Movie). }
    \label{fig:flatGroundExperiments}
\end{figure}

 \begin{figure}
    \centering
    \includegraphics[width=0.7\columnwidth]{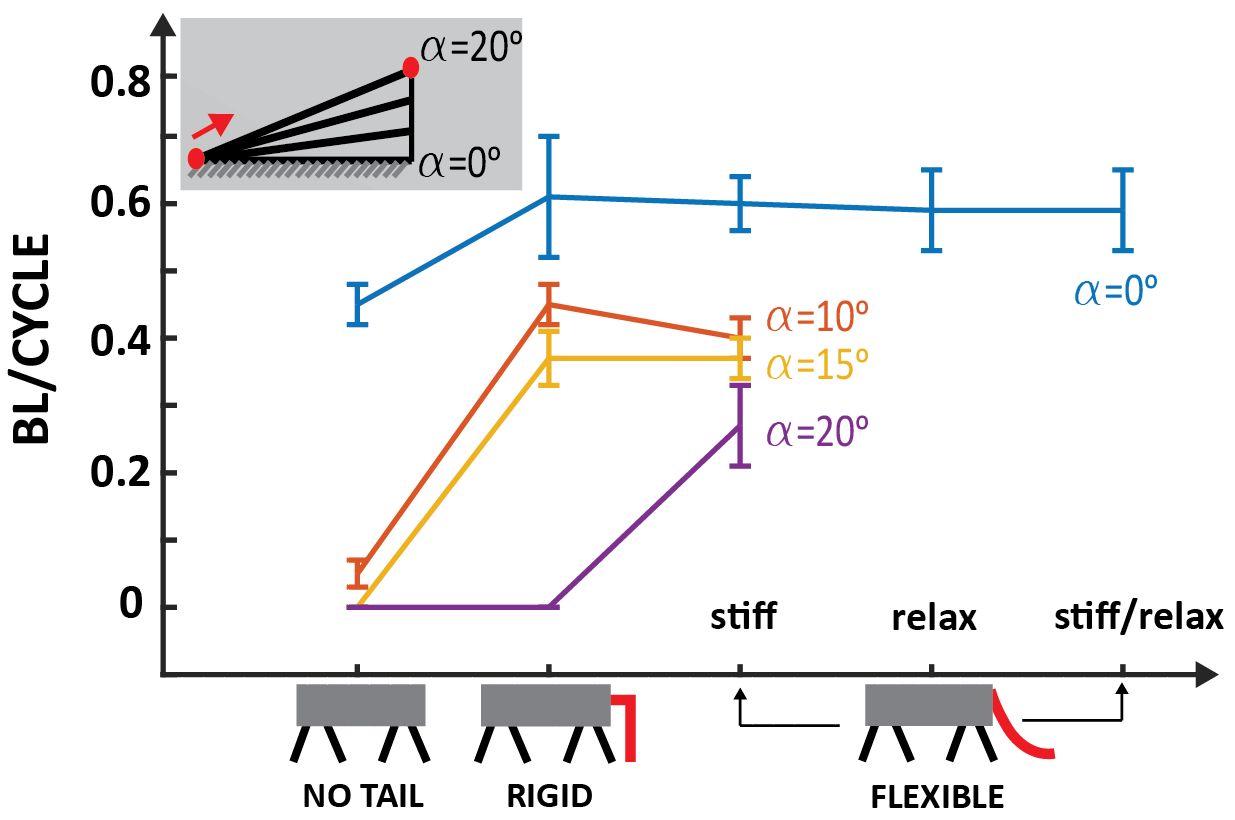}
    \caption{\textbf{Locomotion on a smooth surface with no tail, rigid tail, and flexible tail.} The mean and standard deviation of body length (BL) per cycle for locomotion on smooth flat (blue), $10^o$ (red), $15^o$ (yellow), and $20^o$ (purple) terrain with different tail configurations. Each experiment was repeated at least five times with a minimum of four cycles per run. }
    \label{fig:results1}
\end{figure}

\subsection{Flat Terrain} 
\label{sec:exp_flat_terrain}
The primary objective of the experiments was to evaluate the robot's capacity to maintain a straight trajectory while navigating flat, smooth terrain, which is an essential requirement for various practical applications in the real world. The velocity of the robot was also analyzed for different tail configurations (no tail, rigid tail, and flexible tail), measured in terms of the distance traveled per gait cycle, normalized to the robot's body length (BL/cycle). The experiments aimed to assess the impact of the tail design and properties on the robot's locomotion performance, with particular attention to stability, speed, and accuracy. 

In the experiments, black tape was placed on the surface perpendicular to the ``starting line'' to qualitatively examine the stability of the robot by observing the deviations from the path in each of the configurations. Aspects such as foot clearance and rocking were examined to determine their effect on the robot's balance. Balance disturbances lead to a reduction in locomotion speed, disruption to the gait cycle, and unexpected turning. 

\subsubsection{Locomotion with No Tail}
\label{sec:LocomotionwithNoTail}
This experiment was carried out as described in section \ref{sec:exp_flat_terrain}. In each of the five trials conducted, the robot was unable to meet the goal of traveling along a straight line while on level ground. The robot deviated $2.4\pm0.71^o$ per cycle from the straight path and traveled with a speed of $0.45 \pm 0.03$ BL/cycle (Fig. \ref{fig:results1}). 

This test revealed many problems in the robot's locomotion without a tail. Firstly, the body of the tail struck the ground twice during each gait cycle. The front of the robot pitched upwards during the swing phase causing the rear of the robot to strike the ground before foot placement could occur (Fig. \ref{fig:flatGroundExperiments}A). Secondly, the rear legs of the robot did not have sufficient ground clearance during the swing phase, resulting in the rear swinging leg dragging along the ground. Finally, the front legs were found to have excessive ground clearance, leading to increased motion perpendicular to the ground, rather than the desired motion parallel to the ground. These three problems were inexplicably linked to the upwards pitching of the robot, and caused the robot’s diagonal gait to be asymmetrical and uneven. This impaired the robot's balance, in turn hindering the robot's ability to travel along a straight line. 

\subsubsection{Locomotion with Rigid Tail}
Five replications of the previous experiment were conducted with identical protocols. Quantitative analysis showed a substantial enhancement in locomotion speed ($0.61 \pm 0.09 $), a 34\% increase compared to locomotion with no tail (Fig.\ref{fig:results1}).  However, the robot still could not travel in a  straight line and deviated from the path $3.13\pm1.72^o$ per cycle with a rigid tail.

The tail successfully raised the rear of the robot, eliminating any contact of the body with the ground. Upon close examination of the experiments, it was found that there was still an unpredictable and unwanted variance from the gait cycle. This occurred during the swinging phase of the front right leg. The front right leg made contact with the ground before the back left leg, interrupting the diagonal gait of the robot. This caused the front of the robot to pitch downwards, and the tail to lift off the ground. There was further instability when the tail remade contact with the ground.

\subsubsection{Locomotion with Flexible Tail}
Prior to the experiments being run on the flexible tail, the different tail states/configurations were evaluated. The tail could be configured in the following three ways; \textbf{(1)} a relaxed flexible tail where there was no tension being applied (Fig. \ref{fig:flexibleTailStates}B-left), \textbf{(2)} a more rigid tail where the servo motor was applying constant tension (Fig. \ref{fig:flexibleTailStates}B-middle), or \textbf{(3)} an alternating flexible to rigid tail, which would be controlled by the rotation of the servo motor. 

It was found that the flexible tail with constant rigidity provided the best stability (less vertical oscillation) to the robot while walking on flat, smooth terrain (Fig.\ref{fig:flatGroundExperiments}C). A further improvement in the robot's ability to travel in a straight line was observed in this test. The robot was capable of traveling along a straight path with $0.71\pm0.98^o$ rotation/cycle and $0.60\pm0.04^o$ BL/cycle forward displacement whereas the other tension settings; relaxed and alternating flexible/rigid tail, resulted in $1.48\pm1.77^o$ and $2.5\pm1.77^o$ rotations/cycle. The robot's movements were smoother and more consistent due to the lack of sudden lurches or pitching. It was found that the side-to-side undulations of the tail, although passive and slight, helped maintain ground contact throughout all stages of the gait cycle. The front leg was still making contact with the ground just before the opposite back leg, however, this no longer caused instability. This is because the ground contact of the tail still provided sufficient support to stop rapid pitching.

The presence of residual flexibility in the tail, despite being in a rigid state, emerged as a crucial element in augmenting the robot's stability and locomotion. This allowed the tail to automatically make slight adjustments when traversing the terrain. When the foot strike occurred, the tail moved towards the body momentarily. During the flight phase, the tail then quickly moved in the opposite direction away from the body, until it was limited by the tension of the reel. The subtle movements of the tail played a pivotal role in ensuring that the tail remained in contact with the ground, preserving the overall stability of the robot with seamless transitions between gait phases. The robot's stance position was now altered; the front of the robot was rotated slightly downwards. During the swing phase, the robot pitched marginally upwards until the body angle was approximately parallel to the ground. Then, when the foot strike occurred, the front of the robot pitched downwards to return to the stance position. The swing phase was then again initiated, and the cycle repeated. This resulted in a smooth, consistent gait, without any sudden and rapid pitching taking place, which would cause the robot to change direction.

The constant stability experienced by the robot led to a further increase in locomotion speed. It traveled at a rate of $0.60 \pm 0.04$ BL/cycle similar to the locomotion with the rigid tail.

\subsection{Inclined Surface} In these experiments, we assess the robot's ability to navigate on an inclined surface with each tail configuration. The experiments involved placing a cardboard runway (50 cm x 85 cm) at various inclinations, ranging from 10 to 25 degrees, and testing the robot's ability to traverse the entire length of the runway successfully. Five trials were performed for each incline, and the robot was considered capable of locomotion on an incline if it could make it to the end of the runway successfully (Fig.\ref{fig:results1}).

\subsubsection{Locomotion with No Tail}
With no tail, the robot was unable to successfully travel along any incline. It failed to make it to the end of the runway angled at 10 degrees to the horizontal in five consecutive trials. During each of the trials, pitching occurred, resulting in the body striking the ground as well as low rear leg clearance. The cause of this rocking is identical to what is described in section \ref{sec:LocomotionwithNoTail}.

In the flat terrain tests, the pitching of the robot had only a minor effect on its locomotion capabilities. However, when traveling on an inclined surface, the instability caused by the pitching led to the constant slipping of the robot's rear feet. This made the robot's travel speed extremely slow ($0.05\pm0.02$ BL/cycle). It also resulted in the robot eventually rotating 90 degrees and walking off the runway, therefore failing to meet the success criteria of the experiment.

\subsubsection{Locomotion with Rigid Tail}
The addition of a rigid tail allowed the robot to travel along a runway of inclines 10 and 15 degrees with $0.45\pm0.03$ and $0.37\pm0.04$ BL/cycle, respectively. It is noteworthy that, during locomotion along these inclines, the robot experienced some degree of slippage. However, the hind legs of the robot exhibited adequate ground clearance, thereby facilitating forward movement. When the angle was increased to 20 degrees, the slippage of the feet was greater than the distance stepped forward, therefore, the robot traveled in the reverse direction.

\begin{figure}[t!]
    \centering
    \includegraphics[width=0.7\columnwidth]{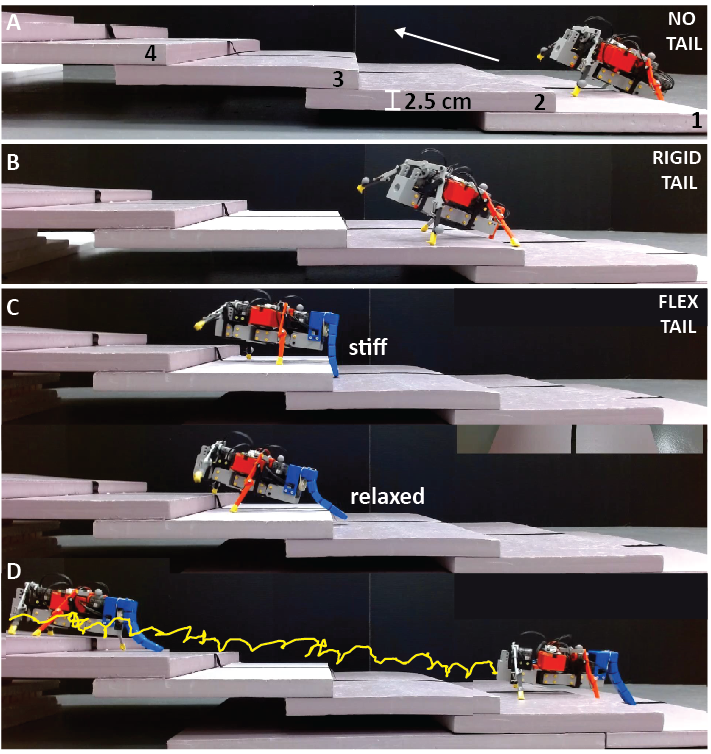}
    \caption{\textbf{Experimental snapshots of the robot climbing up a series of steps. (A)} Without a tail (the robot could not climb onto track), \textbf{(B)} With a rigid tail (robot fell from track at the second step), and \textbf{(C-D)} With a flexible tail that is periodically stiffened and relaxed. The stairs were made of 2.5 cm height foam blocks.  The yellow trajectory shows the trajectory of the tip of the robot during climbing (SI Movie).  }
    \label{fig:locomotionOnStairs}
\end{figure}
\subsubsection{Locomotion with Flexible Tail}
The results obtained from our previous experiments have indicated that the implementation of a flexible tail with constant tension settings yielded the most optimal performance, we therefore conducted our inclined experiments utilizing exclusively this tension setting. When tested with a flexible tail, the robot successfully traveled along a runway of 10, 15, and 20 degrees with $0.40\pm0.03$, $ 0.37\pm0.03$, and $0.27\pm0.06$ BL/cycle, respectively. When the incline of the runway was increased to 25 degrees, the robot could not travel along it due to sliding. The flexibility of the tail was not altered during this experiment and was maintained at high rigidity at all times. This allowed the robot to have constant contact with the runway.

For inclines of 10 and 15 degrees, the robot's motion was only slightly inhibited by some minor slippage of the rear feet. It completed the tests efficiently, while also traveling along the straight line indicated on the runway. In contrast, the robot experienced more pronounced slippage of the front and hind legs when traversing inclines of 20 degrees. Despite this, the robot was able to complete the test, albeit at a slightly reduced speed compared to the preceding trials. However, the robot's performance was severely impaired when the angle of the incline was further increased to 25 degrees. In this case, the robot was unable to make any net progress up the incline, as it slid back to its previous point after each step. It remained here indefinitely, not moving upwards along the track but also not moving downwards and falling off. This suggests that if a change was made to increase the grip of the feet, the robot may be able to successfully move along this incline.

\subsection{Stepped Terrain}  The stepped terrain experiments were split into two sections: \textbf{(1)} upward and \textbf{(2)} downward climbing. Both sections followed identical procedures and had the same success criteria, as detailed below. The only difference between the two sections was that in the upward climb test, the robot initiated from the ground in front of the first step and ascended the steps, while in the downward climb test, the robot commenced from the top step and descended the 'runway' towards the ground, where the ground was considered as the sixth 'step'. 

\begin{figure}
    \centering
    \includegraphics[width=1\columnwidth]{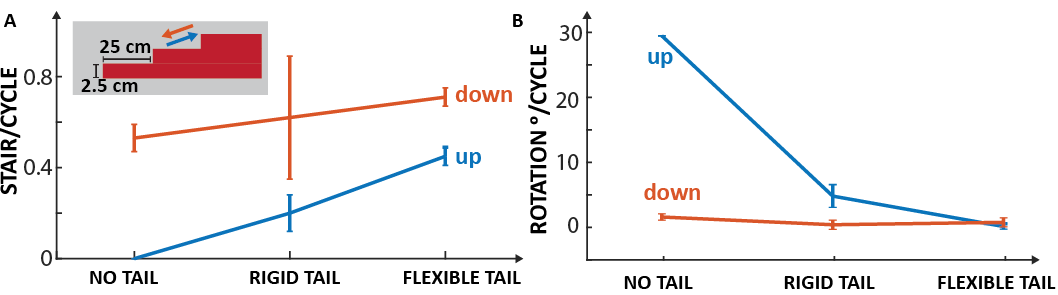}
    \caption{\textbf{Up/down stair climbing with different tail configurations. (A)} Mean and standard deviation of stairs climbed up (blue) and down (red) per cycle without a tail, with a rigid tail, and with a flexible tail over five trials. Inset shows the dimensions of the stairs. \textbf{(B)} Mean and standard deviation of the rotation of the body per cycle for the same experiments.}
    \label{fig:results2}
\end{figure}

 The stair setup comprised six steps of uniform height (2.5 cm) and width (25 cm) made of foam boards (Foamular, HomeDepot). The primary objective was to assess the robot's climbing ability with different tail configurations, with the ultimate goal of climbing as many steps as possible without falling off the runway or failing to make it to the end.  The criterion for a successful climb was that all four feet of the robot should make contact with the steps of the stairs. The total number of steps climbed was then counted over five trials for each tail configuration, with a maximum score of 30 possible, and then normalized by the total number of step cycles. 

\subsubsection{Upward Climb with No Tail}
Without a tail, the robot failed to climb any steps, meaning a result of 0/30, and rotated $30\pm 5^o$ per cycle. It was observed during the previous experiment that on flat terrain the front of the robot pitched up during locomotion, as well as the rear legs' ground clearance was extremely small. These two factors impaired the robot while climbing. During each trial, the robot placed its front legs on the first step. It moved forward on that step until the rear legs reached the edge of the step. When this happened, the rear legs could not lift high enough to clear the ledge. The body was also making contact with the step. As a result, the robot was stuck indefinitely. The result of this experiment was significant, as it showcased the robot's inability to climb or navigate over obstacles without getting stuck. 

\subsubsection{Upward Climb with Rigid Tail}
There was an improvement in the robot's climbing ability when a rigid tail was added. The robot successfully climbed 8/30 steps with $0.2\pm0.08$ stairs/cycle over five trials (Fig.\ref{fig:results2}A) with a $5.34\pm1.75^o$ rotation/cycle. With the addition of the rigid tail, the robot could climb onto the first step in every trial. The robot no longer pitched upwards during locomotion, meaning the hind legs had sufficient ground clearance to make it onto the first step. The robot encountered difficulties when moving forward on the second step and when attempting to climb onto the next step (Fig.\ref{fig:locomotionOnStairs}B). This was due to the lack of contact between the tail and ground, as the robot's legs were on a surface with higher elevation compared to the tail, which led to the loss of support and stabilization. Consequently, the robot's gait reverted to its previous state without a tail, where the front end would incline upwards, and the rear legs would have limited, to no ground clearance. When the tail approached the step's edge, it became stuck, obstructing the robot's progress. Despite being immobilized, the robot continued moving, but without any forward motion along the track. The front legs continued to extend forward, but the tail constrained the robot in place.

In most trials, the robot continued to move forward until, due to slipping of the feet, it began to rotate to one side and ultimately move off the track. In two instances, sudden jerking forward caused by the robot's instability allowed the robot's tail to become free and climb to the next step and continue climbing, although this required numerous gait cycles and a considerable amount of time. It was observed that the robot had already started rotating before being freed, leading to the movement toward the track's edge. Also, when the robot was attempting to climb while on the track, one of the hind legs made insufficient clearance to climb over the step, resulting in additional rotation and ultimately causing the robot to fall off the track. 

It was evident in this test that the length of the robot's tail is a critical factor in determining its climbing ability. It must be noted that although the robot can climb better with a rigid stick-type leg, this rigid leg can easily get caught on obstacles, not just when climbing. It also does not offer sufficient ground support because the contact area of the leg is small. Increasing the tail's length increases the clearance height of the rear of the robot and provides more support during climbing, which can enhance the robot's stability and balance. However, longer tails are more likely to get caught on objects or ledges, leading to potential failures during locomotion. Meanwhile, shorter tails decrease the probability of getting stuck but also decrease the support provided during climbing and flat ground traversal, leading to reduced stability and insufficient ground clearance.

\subsubsection{Upward Climb with Flexible Tail}
Making the tail longer and more flexible allows the robot to adapt the height of its backside to the environment and provide more support during climbing. Since the robot with the rigid tail could not climb the stairs, here we programmed the flexible tail to periodically stiffen and relax during key phases of the gait cycle. Tail stiffening occurred during the middle of the flight phase of the front left leg i.e. when it was reaching forward. Tail relaxation occurred at the same instant during the flight phase of the front right leg. The robot successfully climbed 30 steps out of a possible 30, with $0.45\pm 0.04$ stairs/cycle and $0.66\pm0.41^o$ rotation/cycle, over 5 trials (Fig.\ref{fig:results2}). 

When the tail is stiffened, it allows the robot to climb to a higher step by providing support in the form of constant ground contact, resulting in a large clearance distance between the legs and the ground. While the robot is on the higher step, the body remains parallel, resulting in smoother locomotion. However, due to the length of the tail, it can get stuck when it reaches the edge of a step, and the robot ceases to move forward.

To prevent this, the tail relaxes by releasing the tension on the cables. This allows the tail to move over the step and for the robot to continue climbing. Due to the periodic nature of the stiffening and relaxation of the tail, in some cases, the tail may inhibit movement and the robot may appear stuck momentarily. This occurs when the tail, in its stiffened state, meets a step. The robot then attempts to move forward for a brief period. During the forward swing, the tail is relaxed, and the robot accelerates rapidly forward. The feet of the robot slides along the surface and as a result, the robot moves a further distance forward than a typical cycle. The stiffening of the tail also provides thrust to the robot, pushing it upward and forward at a specific instant during the gait cycle, to increase locomotion effectiveness.  

\subsubsection{Downward Climb with No Tail}
When the robot has no tail attached, it performs better at climbing downwards compared to climbing upwards. The robot successfully traveled down 15 steps out of a possible 30 with $0.53\pm0.06$ stairs/cycle and $2.12\pm0.47^o$ rotation/cycle, over six trials. From previous experiments it was observed, that when traveling along flat terrain the robot pitches upwards, causing the rear to make contact with the ground. This still occurred when climbing downwards, but when the robot moved to a lower step, it violently pitched downwards, causing the front to hit the ground or runway. This caused further disturbance to the robot’s gait, leading to instability and inconsistencies in the robot’s direction of travel.	

\subsubsection{Downward Climb with Rigid Tail}
The robot successfully traveled down 21 steps with $0.62\pm0.27$ stairs/cycle and $0.92\pm0.71^o$ rotation/cycle when a rigid tail was attached. The robot’s locomotion patterns were made more consistent by the addition of a tail. However, upon each foot placement, the robot pitched downwards and the tail briefly lost contact with the supporting surface. This caused problems when moving onto a lower step. In those cases, the robot pitched to an even more extreme angle, resulting in the front of the robot making contact with the surface. Similar to the previous experiment, these collisions with the runway resulted in negative consequences such as misdirection and less efficient locomotion, albeit to a lesser degree.

\subsubsection{Downward Climb with Flexible Tail}
With the addition of the flexible tail, the robot climbed down 30 out of a possible 30 steps with $0.71\pm0.04$ stairs/cycle and $1.3\pm0.7^o$ rotation/cycle. Different to the upward climbing where the tail was periodically stiffened and relaxed, in these experiments the tail was configured so that it was fully flexible i.e., there was no tension in the cables and there was no periodic motion of the reel. This configuration of the tail made the robot more adaptable to its terrain. When stepping off each step, the tail extends to the point where it is almost entirely vertical. At this instant, the bottom of the tail is in direct contact with the ground (the narrow end piece of the tail). Simultaneously, the front leg springs bend, and the front of the robot makes contact with the ground. However, as the tail is still supporting the robot, there is the little impact felt by it and there are no deviations from the path. At all other stages, the tail is flexed away from the body of the robot. This allows the tail to continue to offer support when the legs of the robot have moved to a lower step. 

\subsection{OUTSIDE EXPERIMENTS} 
To demonstrate the tail function on locomotion, outside experiments were carried out on the following terrains: a pebbled surface, mulch ground with plant litter (twigs, leaves, etc.), and sand. The key parameter to be analyzed in this experiment was how the robot reacted to uneven terrains, and if the addition of a flexible tail improved these interactions This was done qualitatively. In this section, tests were only performed on the robot with no tail and with a flexible tail.

\begin{figure}[h!]
\centering
    \includegraphics[width=0.97\textwidth]{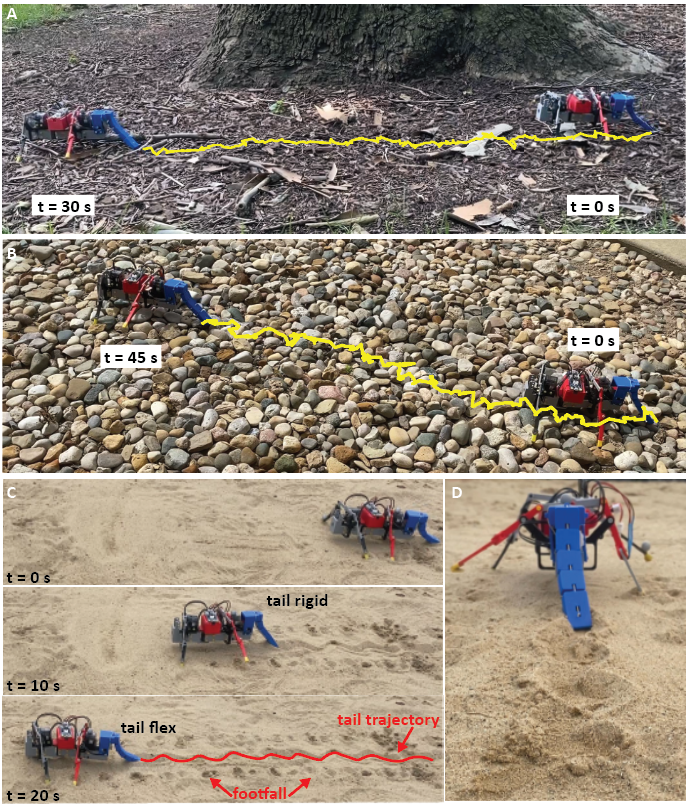}
    \caption{\textbf{Outside demonstrations.} \textbf{(A)} Side view of the robot while walking on mulch surface. Yellow trajectory shows the trajectory of the tip of the tail during walking (t=0-30 s), \textbf{(B)} Locomotion on a pebbled surface. \textbf{(C)} Side view of the robot while walking on sand. At t=0 and 20s the tail of the robot is in its relaxed state, and at t=10s it is in its rigid state. The red trajectory shows the tail trajectory. Also, the footfalls of the robot are highlighted with red arrows, \textbf{(D)} Back view of the robot and its tail trajectory while walking on the sand (SI Movie). }
    \label{fig:outsideExperiments}
\end{figure}

\subsubsection{Mulch Terrain}
Without a tail, the robot was able to traverse on mulch without being significantly affected by the softness of the surface (Fig.\ref{fig:outsideExperiments}A).  However, when the robot encountered organic debris like leaves and fallen twigs,  the legs of the robot would drag the debris along, resulting in a reduction to the robot's velocity. When a flexible tail was added, which exhibited periodic stiffening and relaxation, the robot was more resilient to the aforementioned obstacles (SI movie). The tail's flexibility allowed it to absorb the impacts of the obstacles and facilitate the robot's smooth movement.

\subsubsection{Pebbled Terrain}
With no tail, the robot traveled extremely slowly ($\sim$ zero speed) over pebbled surfaces. The robot's legs repeatedly got stuck in gaps between stones, due to the low ground clearance of the legs and the pitching of the body. This was compounded by the robot's eventual entrapment, which resulted from one of its legs becoming firmly lodged between stones, rendering it immobile. 

When the robot was tested with a flexible tail that exhibited periodic stiffening and relaxation, the locomotion velocity was increased to $0.26\pm0.04$ BL/cycle (SI movie). The legs no longer got stuck due to the movement of the tail and the support it provided. Although the tail occasionally became stuck, the periodic relaxation of the tail allowed it to become dislodged quickly without impeding the robot's overall performance (Fig.\ref{fig:outsideExperiments}B).

\subsubsection{Sand}
When traveling on sand, the robot's rear legs did not lift off the surface as expected, but instead were dragged along through the sand. This unexpected movement of the robot's legs resulted in disruptions to the gait cycle and subjected the rear legs to increased force and twisting, which the robot was not designed to handle (SI movie). Continued use of the robot on sand would likely lead to fracturing of the legs or body, as well as damage to the rear motor due to the excessive forces being exerted. 

In contrast to the issues encountered when the robot traversed a sandy terrain without a tail, the addition of a flexible tail configured to periodically stiffen and relax proved to be an effective solution (Fig.\ref{fig:outsideExperiments}C). The tail's ability to provide constant support and its side-to-side passive undulation was clearly visible in the tracks left by the robot(Fig.\ref{fig:outsideExperiments}C-D). Specifically, the tail enabled the robot to traverse small divots and uneven surfaces, which it was previously unable to navigate effectively (SI movie). 

\section{Conclusions and Future Works}
This paper describes the development and implementation of a multi-segment, flexible tail with variable rigidity controlled by a cable-driven mechanism. The performance of a sprawling quadruped robot was shown to be significantly improved by conducting experiments that compared the locomotion of the robot with different tail configurations, over several indoor and outdoor environments.

The constant ground support provided by the flexible tail is key in maintaining stable locomotion. This ensures a predictable gait cycle that stops unexpected turning and slipping as experienced by the robot in the other two tail configurations. This leads to an increase in the locomotion speed and efficiency of the robot. Moreover, the constant ground support also enables the robot to travel along an inclined surface with an angle of up to 20 degrees to the horizontal. The variable stiffness of the tail plays a significant role in enhancing the robot's climbing ability. The versatility and adaptability of the tail allows the robot to successfully overcome obstacles in its path when climbing, without the need for external sensing.

All of the aforementioned improvements to the robot's locomotion were observed when testing on natural terrain occurred. The adaptability of the flexible tail greatly improved the robot's locomotion capabilities when traveling on pebbled, mulch, and sandy terrain.

A limitation of the open-loop system employed in conjunction with the flexible tail is its inability to change the state of the flexible tail automatically. The configuring of the tail i.e. full flexibility, constant rigidity, or periodic rigidity, must take place before the beginning of the experiments, based on the terrain. However, future work will focus on the integration of various sensors to automatically change the state of the tail according to the terrain properties. This will allow the robot to adapt to its terrain and change its gait to effectively navigate obstacles, with the goal of applying the robot autonomously in outdoor, real-world environments. Our preliminary testing of the incorporation of touch sensors at the end of the tail has shown promising results in aiding the robot during stair climbing, with the relaxing and stiffening of the tail determined by the sensor (SI Movie). 

With the implementation of a closed-loop system, the robot would travel with the tail in its rigid state, when the surface is smooth or flat. If an obstacle is sensed in front of the robot, the tail would periodically relax and stiffen to prevent entrapment. Similarly, if the surface underneath the robot were to fall off, the tail would fully relax, providing support while the robot pitched downwards. These advances may lead to the autonomous application of the legged robot in outdoor, real-world environments, where it can be used for various tasks, such as search and rescue, environmental monitoring, and inspection.

\section*{Author Contributions}

JB designed the different tail configurations, built the robot, and performed the experiments, YOA designed and supervised the project, and both authors analyzed the data and wrote the paper.

\section*{Funding}
Josh Buckley is supported by The Naughton Undergraduate Fellowship program.

\section*{Acknowledgments}
We would like to thank Callie King for her assistance in the assembly of the robot body and stair setup. We would also like to express our appreciation to the members of Notre Dame MiNiRo-Lab for their insightful discussions.

\section*{Supplemental Data}
The supplementary movie can be found in the submission files.

\section*{Conflict of interest}
The authors declare that the research was conducted in the absence of any commercial or financial relationships that could be construed as a potential conflict of interest.

\bibliographystyle{plain} 

\end{document}